\documentclass{article}
\usepackage{acra}

\usepackage{booktabs}
\usepackage{amsmath}
\usepackage{graphicx} 
\usepackage{xcolor}
\definecolor{darkred}{RGB}{179,0,27}
\usepackage[colorlinks=true, allcolors=darkred]{hyperref}

\newcommand{\bs}{\beta_{\mathrm{s}}}
\newcommand{\be}{\beta_{\mathrm{e}}}
\newcommand{\reach}{\texttt{Fetch-Reach}}
\newcommand{\push}{\texttt{Fetch-Push}}
\newcommand{\pick}{\texttt{Fetch-PickAndPlace}}
\newcommand{\rotate}{\texttt{ShadowHand-BlockRotateZ}}

\usepackage{fancyhdr}
\fancypagestyle{firstpage}{
    \fancyhf{} 
    \fancyhead[C]{\footnotesize Cite as: ``López, F. M., and Cruz, F. (2026). Fade to... Red? Scheduling Coloured Noise Exploration Supports Efficient Robot Learning with Reduced Movement Jerk. In Proceedings of the Australasian Conference on Robotics and Automation (ACRA). Australian Robotics and Automation Association (ARAA).''} 
    \fancyfoot[C]{\thepage} 
}

\title{
Fade to\dots Red? Scheduling Coloured Noise Exploration Supports Efficient Robot Learning with Reduced Movement Jerk
}

\author{Francisco M. López$^{1,2}$ and Francisco Cruz$^{1,3}$ \\
$^1$ School of Computer Science and Engineering, UNSW Sydney, Australia\\
$^2$ Frankfurt Institute for Advanced Studies, Germany \\
$^3$ Escuela de Ingeniería, Universidad Central de Chile, Chile \\
\{fm.lopez, f.cruz\}@unsw.edu.au
}

\begin{document}

\maketitle

\thispagestyle{firstpage}

\begin{abstract}
Exploration in deep reinforcement learning (RL) for continuous control is commonly driven by temporally uncorrelated Gaussian (white) action noise, which leads to jerky motions with locally constrained exploration. In contrast, temporally correlated coloured noise, particularly pink noise, has been shown to support more efficient exploration in RL. Motivated by these findings, we inquire whether scheduling changes to the noise colour over the course of training can outperform white noise for continuous control robotics. We perform a systematic study of 49 linear schedules between white and red noise, together with generalised state-dependent exploration (gSDE), on four sparse-reward manipulation tasks with a Fetch arm and a Shadow Dexterous Hand. We find that the noise colour at the end of training does not affect learning efficiency or end performance, whereas fading towards red noise halves the jerk of exploration in every task. Balancing learning efficiency, end performance and smoothness, red noise emerges as the preferred strategy: it learns as efficiently as pink noise with almost 2-fold less movement jerk. We therefore recommend exploration schedulers that fade towards red noise as the default for off-policy robot learning.
\end{abstract}

\section{Introduction}
\label{sec:intro}

\begin{figure*}
    \centering
    \includegraphics[width=1\linewidth]{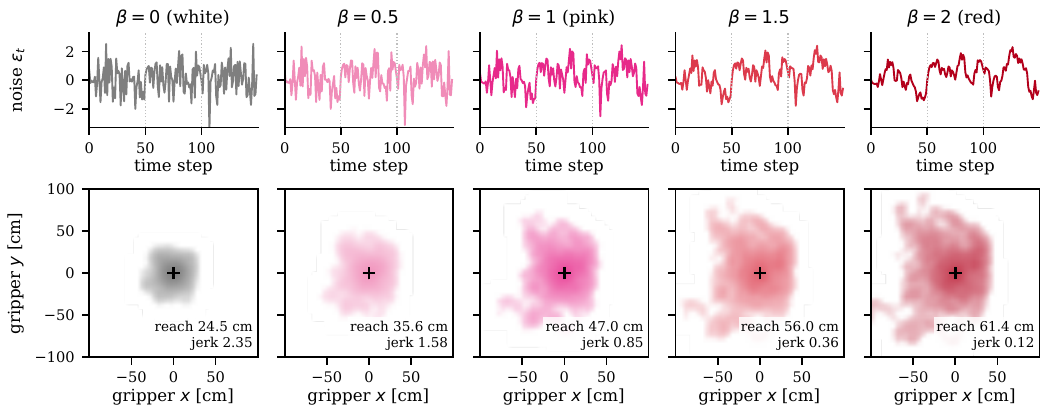}
    \caption{Coloured noise in a Fetch arm. \textbf{Top}: Examples of noise sequences with noise colours ranging from white ($\beta=0$) to red ($\beta=2$). All sequences were generated from the same initial Gaussian white noise sequence with unit variance. \textbf{Bottom}: Densities of gripper positions relative to the start position (black cross) over 100 episodes of 50 steps with the arm driven by exploration noise alone. Inset values show the maximal gripper displacement and average action jerk over those 100 episodes. These illustrative examples show that increasingly correlated noise generated from the same initial Gaussian white noise reaches a wider area with smoother movements.}
    \label{fig:fetch_noise}
\end{figure*}

Reinforcement learning (RL) offers a path for robots to acquire skills through trial and error~\cite{kober2013survey,ibarz2021train}. In continuous action spaces, learning requires exploration through action noise: random perturbations of the policy that allow the robot to discover new behaviours. Off-policy algorithms such as Soft Actor-Critic (SAC) \cite{haarnoja2018sac} sample perturbations independently at every time step, i.e. they explore with temporally uncorrelated Gaussian \emph{white} noise. In the context of robotics, white noise has two critical drawbacks. First, exploration is local. \cite{eberhard2023pink} showed that white noise perturbations cancel out over time, so the robot's trajectories only rarely reach areas far from its learned policy, thus limiting the robot from encountering sparse rewards (see Fig.~\ref{fig:fetch_noise}unsafe). Second, white noise produces jerky, high-frequency motor commands that are undesirable on physical hardware, where they cause vibrations and actuator wear and can be unsafe~\cite{raffin2022gsde,ibarz2021train}.

These two drawbacks can be simultaneously addressed by using temporally correlated coloured noise for exploration. Coloured noise is characterised by a power spectral density (PSD) that follows $S(f) \propto f^{-\beta}$. Larger values of $\beta$ result in higher presence of low frequencies relative to high frequencies, i.e. more temporal autocorrelation and thus smoother movements. White noise is a particular case with $\beta = 0$. Other common coloured noises include pink noise ($\beta = 1$) and red (Brownian-like) noise ($\beta = 2$). Recent work has shown that intermediate correlations achieve the highest average scores on common RL environments \cite{eberhard2023pink,pinneri2021icem,lopez2026baby}. In light of these results, pink noise is now deemed the recommended default action noise for continuous control.

In this paper, we explore the problem of exploration noise for robotic manipulation. Following on the results of \cite{lopez2026baby}, we inquire whether a schedule that changes the noise colour during training may be beneficial in the context of robotics. Our contributions are as follows:
\begin{itemize}
  \item A systematic study of linear noise-colour schedules covering all $7 \times 7$ combinations of start and end colours $\beta \in \{0, 0.5, 0.75, 1, 1.25, 1.5, 2\}$, whose diagonal contains the constant colours, plus generalised state-dependent exploration (gSDE)~\cite{raffin2022gsde}, on four sparse-reward manipulation tasks.
  \item An evaluation of learning efficiency, end performance, and smoothness of the exploratory motions that a physical robot would execute during training.
  \item The finding that scheduling decouples learning from smoothness: the noise colour late in training does not affect learning efficiency or end performance, whereas fading towards red noise halves the jerk of exploration in every task.
  \item A recommendation for robot learning derived from the trade-offs between the three metrics: red exploration noise is statistically tied with the best strategies in learning efficiency and end performance and produces the smoothest exploration of all strategies.
\end{itemize}

\section{Related Work}
\label{sec:related}

\paragraph{Action noise for exploration.}
Exploration remains a central challenge in RL~\cite{amin2021survey}. In continuous control, the simplest and most widely used strategy is to perturb actions with noise. Deterministic-policy algorithms such as DDPG~\cite{lillicrap2016ddpg} and TD3~\cite{fujimoto2018td3} add Gaussian white noise or Ornstein-Uhlenbeck (OU) \cite{uhlenbeck1930brownian} noise to the policy output, whereas stochastic-policy algorithms such as SAC~\cite{haarnoja2018sac} sample actions from a state-dependent Gaussian, which amounts to scale-modulated white noise~\cite{eberhard2023pink}. \cite{hollenstein2022action} showed that the best noise type and scale depend on the environment. Alternatives to action-space noise perturb the policy parameters~\cite{plappert2018parameter} or repeat exploratory actions over random durations~\cite{dabney2021temporally} with embodiment-specific correlations for overactuated systems~\cite{schumacher2023dep}.

\paragraph{Coloured noise.}
Coloured noise leads to efficient exploration in model-predictive control~\cite{pinneri2021icem} and model-free RL~\cite{eberhard2023pink}. Coloured noise is characterised by a noise colour parameter $\beta$, the negative slope of the PSD $S(f) \propto f^{-\beta}$. The noise colour $\beta=1$, referred to as pink noise, performs on par with an oracle that selects the best constant colour per task \cite{eberhard2023pink}. Its benefits extend to on-policy algorithms such as PPO~\cite{hollenstein2024ppo}, and its effect on the optimal policy has been analysed in linear-quadratic settings~\cite{hollenstein2024lqr}. All these experiments were performed on standard RL benchmarks such as the DeepMind Control Suite and MuJoCo locomotion. However, robotic manipulation tasks and their hardware-relevant constraints have received little attention.

\paragraph{Scheduled and adaptive exploration.}
Annealing the amount of exploration is common practice in RL, from decaying $\epsilon$-greedy schedules to reducing the scale of action noise during training, which can improve the performance of off-policy agents~\cite{hollenstein2022action}. Recent work has shown that a developmentally-inspired scheduled noise that recreates the changes in spontaneous movements of human infants by increasing the noise colour from 0.7 to 0.9 during training leads to marginally better performance than any fixed coloured noise \cite{lopez2026baby}. However, it is unclear whether this particular scheduled coloured noise presents any advantage over any other scheduled exploration strategy. To the best of our knowledge, no previous study has systematically compared the start and end colours of noise schedules.

\paragraph{Smooth exploration for robot learning.}
On physical robots, the smoothness of exploratory actions matters as much as their efficiency~\cite{ibarz2021train}. gSDE~\cite{raffin2022gsde} replaces per-step noise with a state-dependent perturbation whose parameters are resampled only occasionally, yielding smooth exploration that enabled learning directly on real robots. Other works have explored the regularisation of the smoothness of the learned policy itself~\cite{mysore2021caps}. Coloured noise offers a complementary way of controlling the smoothness of exploration through a single parameter: the noise colour $\beta$.

\section{Scheduled Coloured Noise}
\label{sec:method}

\subsection{Exploration noise in policy learning}

Coloured noise can be implemented in both deterministic-policy algorithms, e.g. DDPG or TD3, and stochastic-policy algorithms, e.g. SAC~\cite{haarnoja2018sac}. This paper focuses exclusively on SAC, since it is the more widely used algorithm for robotics. However, the same formalism can generalise to TD3 (see \cite{lopez2026baby}). 
SAC learns a stochastic policy that maximises the expected return plus the entropy of the policy. Actions are sampled from a squashed Gaussian using the reparameterisation
\begin{equation}
  a_t = \tanh\!\big(\mu_\theta(s_t) + \sigma_\theta(s_t) \odot \varepsilon_t\big), \quad \varepsilon_t \sim \mathcal{N}(0, I),
  \label{eq:sac}
\end{equation}
where $\mu_\theta$ and $\sigma_\theta$ are outputs of the policy network. Because $\varepsilon_t$ is drawn independently at every step, the exploration noise $\varepsilon_{1:T}$ is white noise. SAC is off-policy, so the noise used to collect experience can be replaced by any other process without needing to change the learning objective~\cite{eberhard2023pink,lopez2026baby}.

A stationary Gaussian process is called coloured noise with noise colour $\beta$ if its PSD follows $S(f) \propto f^{-\beta}$. We can generate it by spectral shaping~\cite{timmer1995generating} of white noise: Gaussian random Fourier coefficients are scaled by $f^{-\beta/2}$ and transformed to the time domain with an inverse Fourier transform, and the result is normalised to unit variance. Coloured noise therefore has the same standard normal marginal distribution as white noise but differs in its temporal correlation~\cite{eberhard2023pink}. Fig.~\ref{fig:fetch_noise} shows examples of different coloured noises. The white noise $\varepsilon_t$ in~\eqref{eq:sac} can be replaced by a coloured noise to promote more efficient exploration and learning \cite{eberhard2023pink}.

\subsection{Linear colour schedules}

It is also possible to use a scheduled coloured noise where the noise colour $\beta$ is annealed during training \cite{lopez2026baby}. Given a start colour $\bs$, an end colour $\be$ and a training budget of $T$ environment steps, the colour at step $t$ is
\begin{equation}
  \beta(t) = \bs + (\be - \bs)\,\frac{k(t)}{K - 1}, \qquad k(t) = \Big\lfloor \frac{t K}{T} \Big\rfloor,
  \label{eq:schedule}
\end{equation}
where the training is divided into $K$ blocks of equal length within which the colour is constant, so that noise can be generated in long sequences (see \cite{lopez2026baby}). We use $K = 20$ for all tasks, so that the schedule resolution does not depend on the budget. The first block uses $\bs$ and the last block $\be$. Constant noise corresponds to $\bs = \be$. Independent noise sequences are generated for each action dimension.

In common implementations of SAC, including the Stable-Baselines3 implementation used here \cite{raffin2021sb3}, the same noise sampling routine is used for environment interactions and gradient updates. Our study focuses exclusively on the role of coloured action noise for exploration. Therefore, we apply coloured noise only during interactions of the robot agent with the environment whereas gradient updates use the standard i.i.d. Gaussian reparameterisation.

\section{Experimental Setup}
\label{sec:setup}

\subsection{Tasks}
We use four goal-conditioned manipulation tasks from Gymnasium-Robotics~\cite{gymnasiumrobotics2023,plappert2018multigoal}, simulated in MuJoCo~\cite{todorov2012mujoco}. Three tasks use a 7-DoF Fetch arm controlled through an action space of dimension $|\mathcal{A}|=4$ (three Cartesian displacement of the gripper plus the gripper opening): moving the gripper to a target position (\reach), pushing a block to a target on the table (\push), and grasping a block and placing it at a target position (\pick). One additional task uses the 24-DoF Shadow Dexterous Hand~\cite{openai2020dexterous} with 20 position-controlled actuators to rotate a block in the palm to a target orientation about the $z$ axis (\rotate). All tasks have sparse rewards and report success when the goal is achieved within a tolerance. 

\subsection{Exploration strategies}

We evaluate all 49 linear schedules with noise colours $\bs, \be \in \{0, 0.5, 0.75, 1, 1.25, 1.5, 2\}$, covering white ($\beta = 0$), pink ($\beta=1$) and red ($\beta=2$) noise as well as intermediate noise colours. The seven schedules with $\bs = \be$ are the constant noise colours, including the SAC default (white) and the recommended default (pink) proposed by \cite{eberhard2023pink}. The grid contains both increasing and decreasing schedules. As a robotics-oriented baseline we also one additional exploration strategy: gSDE~\cite{raffin2022gsde}, with the exploration matrix resampled at the start of every episode.

\subsection{Training}

Since the tasks considered here are goal-conditioned with sparse rewards (the agent receives $-1$ at every step until the goal is achieved~\cite{plappert2018multigoal}), we use Hindsight Experience Replay (HER)~\cite{andrychowicz2017her} to relabel stored transitions with goals that were actually achieved later in the same episode, turning failed episodes into useful training data. All agents use the SAC implementation of Stable-Baselines3~\cite{raffin2021sb3} and HER with four relabelled goals per transition. Training hyperparameters are kept identical across all exploration strategies. The actor and critic networks both consist of two hidden layers of 256 units with ReLU activations. The networks are optimised using Adam with a learning rate of $10^{-3}$. We use a replay buffer capacity of $10^6$ transitions, a batch size of 256, a target smoothing coefficient $\tau$ of 0.05, and an automatic entropy coefficient with a target of $-|\mathcal{A}|$ (action space dimension). We use $K=20$ scheduled colour noise blocks and a noise cutoff of $f_{\min} = 1 / T_{\mathrm{ep}}$.

Some hyperparameters are varied across tasks according to their demands, with their values determined from a set of pilot runs of 10 seeds with white noise. All Fetch tasks have a maximum episode length $T_ep$ of 50 steps and use a discount factor $\gamma=0.95$. The \rotate\ has $T_ep=100$ and $\gamma=0.98$. The total training budget (in total environment steps) is 20,000 for \reach; 500,000 for \push; and 1,000,000 for both \pick\ and \rotate. 

The first 1,000 steps use uniformly random actions to initialise the replay buffer. The scheduled coloured noise is applied from that point until the end of training. Agents collect one episode worth of transitions and then perform as many gradient steps as environment steps collected, which is standard for HER and allows gSDE to keep its exploration matrix fixed within an episode. Each of the $4~\times~50~=~200$ task-noise pairs is trained with 30 random seeds, giving a total of 6,000 runs. Seeds are shared across strategies, and the noise generators use the same seeds, so that runs of different strategies are paired through common random numbers~\cite{lopez2026baby}.

\subsection{Metrics}
\label{sec:metrics}

\paragraph{Learning efficiency and end performance.}
Every $T/50$ steps we evaluate the deterministic policy ($a_t = \tanh \mu_\theta(s_t)$) on 20 episodes and record the success rate. Our main performance measure is the area under the learning curve (AUC), i.e.\ the mean success rate over the 50 evaluations, which captures both learning speed and final performance~\cite{eberhard2023pink}. We also report the final success rate (end performance) as the mean success over the last 5 evaluations, i.e. the last 10\% of training.

\paragraph{Smoothness and effort of exploration.}

For every training episode we record the executed actions $a_{1:T_{\mathrm{ep}}}$ and compute the action jerk:
\begin{equation} \label{eq:jerk}
    j = \frac{1}{(T_{\mathrm{ep}}-2)|\mathcal{A}|}\sum_t \lVert a_t - 2a_{t-1} + a_{t-2} \rVert^2.
\end{equation}

For the Shadow Hand, we additionally record the effort as the mean squared actuator force. These metrics are averaged over all training episodes of a run.

\paragraph{Standardisation relative to white noise.}

To aggregate results across all four different tasks, which have different difficulties and training budgets, we express all metrics as z-scores relative to white noise, the default exploration noise of SAC. For the learning efficiency and end performance, we subtract the mean and divide by the standard deviation of the 30 white-noise runs. Therefore, positive metrics indicate better performance than the default white noise and negative ones indicate worse performance. We also compute a jerk ratio for each exploration strategy as the ratio of its absolute jerk, defined by \eqref{eq:jerk}, to the absolute jerk of white noise.

\begin{figure*}[!t]
    \centering
    \includegraphics[width=1\linewidth]{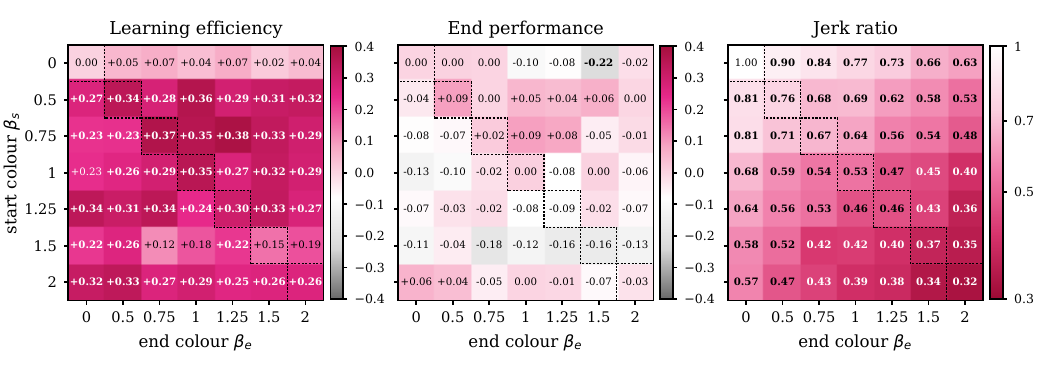}
    \caption{Aggregated results from all four tasks and 30 seeds for each of the 49 coloured noise exploration strategies. \textbf{Left}: learning efficiency (AUC). \textbf{Center}: End performance from last 10\% of training. \textbf{Right}: action jerk during training. All values are normalised with respect to the white noise performance. Values in \textbf{bold} indicate significant differences to white noise on 95\% paired bootstrap confidence intervals. The gSDE scores (not shown) are $-0.75$, $-0.52$, and $0.44$, respectively. }
    \label{fig:grid}
\end{figure*}

\subsection{Statistical analysis}

95\% confidence intervals are obtained with a paired bootstrap that resamples seeds within each task while keeping all strategies of a seed together, thus preserving the pairing through common random numbers (4,000 resamples). To determine if and when noise colour matters, we separate the effects of start and end colours in three ways. First, we compare end colours within groups of runs that share seed and start colour (paired contrasts and Friedman tests). Second, we compare start colours after averaging over end colours within each seed (Friedman tests with seeds as blocks, and Spearman correlations). Third, we regress the z-scored metrics on the mean colour $\bar\beta = (\bs + \be)/2$, its square, and the colour change $\Delta\beta = \be - \bs$, with task fixed effects and a bootstrap over seeds. Finally, we compare each of the non-constant schedules with the constant noise of the same start colour (Wilcoxon signed-rank tests over seeds and tasks, Bonferroni-corrected).

\subsection{Pareto frontiers}

To balance learning, end performance and smoothness, we identify the Pareto frontiers of the 50 different exploration strategies in the planes of learning efficiency vs.\ jerk ratio and end performance vs.\ jerk ratio, i.e.\ the strategies for which no other strategy is at least as good in both metrics and better in one. Because differences in learning between many strategies are small relative to the variability between seeds, we establish an optimality decision rule: among the strategies whose learning efficiency (or end performance) is not significantly worse than that of the best strategy (paired bootstrap), the optimal exploration strategy is the one with the lowest jerk. This decision rule is most relevant for embodied robots, where high movement jerk can be problematic.

\section{Results}
\label{sec:results}

\subsection{How noise colour shapes exploration}

To illustrate the effects of noise colour on exploration, we drive the Fetch arm with pure noise, i.e.\ the actions of an untrained SAC policy ($\mu_\theta = 0$, $\sigma_\theta = 1$ in~\eqref{eq:sac}) for 50-step episodes. White noise keeps the gripper close to its starting position, whereas increasingly coloured noise produces longer, smoother excursions. Over 100 episodes per colour, the mean maximal gripper displacement grows from 24.5\,cm with white noise to 47.0\,cm with pink and 61.4\,cm with red noise, while the action jerk falls 19-fold, from 2.35 to 0.12. More correlated noise therefore explores further and more smoothly. However, it also produces data that is increasingly off-policy and can drive the robot into the boundaries of its workspace \cite{eberhard2023pink,lopez2026baby}. To avoid this negative consequence of temporally correlated noise, \cite{eberhard2023pink} suggest using pink noise, i.e. a moderately correlated noise, by default. In contrast, we suggest that scheduled coloured noise that changes the balance between local and global exploration as learning progresses may be a more suitable solution for robotics.

\begin{table*}[!t]
  \centering
  \caption{Aggregate learning efficiency and end performance (z-scores relative to white noise, mean over tasks) and jerk ratio relative to white noise (geometric mean over tasks) for some representative exploration strategies. Brackets indicate 95\% paired bootstrap confidence intervals over 30 seeds. The best values in each column are highlighted in \textbf{bold} and the second best values are \underline{underlined}.}
  \label{tab:main}
  \small
  \resizebox{\textwidth}{!}{%
  \begin{tabular}{@{}lccc@{}}
    \toprule
    \textbf{Exploration strategy} & \textbf{Learning efficiency ($\uparrow$)} & \textbf{End performance ($\uparrow$)} & \textbf{Jerk ratio} ($\downarrow$) \\
    \midrule
    \textit{Common baselines} \\
    White noise ($\beta=0$)                  & $0.00$ \scriptsize{$[-0.18, +0.19]$} & $0.00$ \scriptsize{$[-0.18, +0.17]$} & $1.00$ \\
    gSDE                               & $-0.75$ \scriptsize{$[-0.88, -0.63]$} & $-0.52$ \scriptsize{$[-0.74, -0.33]$} & $0.44$ \scriptsize{$[0.38, 0.50]$} \\
    \midrule
    \textit{Fixed coloured noise} \\
    $\beta=0.5$              & $+0.34$ \scriptsize{$[+0.14, +0.55]$} & \underline{$+0.09$} \scriptsize{$[-0.08, +0.27]$} & $0.76$ \scriptsize{$[0.69, 0.84]$} \\
    $\beta=1$ (Pink noise)                   & $+0.35$ \scriptsize{$[+0.18, +0.52]$} & $0.00$ \scriptsize{$[-0.17, +0.17]$} & $0.53$ \scriptsize{$[0.47, 0.60]$} \\
    $\beta=2$ (Red noise)                    & $+0.26$ \scriptsize{$[+0.11, +0.42]$} & $-0.03$ \scriptsize{$[-0.18, +0.13]$} & $\mathbf{0.32}$ \scriptsize{$[0.28, 0.37]$} \\
    \midrule
    \textit{Scheduled coloured noise} ($\bs \to \be$)  \\
    $0 \to 1$ (White to Pink)       & $+0.04$ \scriptsize{$[-0.13, +0.22]$} & $-0.10$ \scriptsize{$[-0.30, +0.09]$} & $0.77$ \scriptsize{$[0.72, 0.83]$} \\
    $0 \to 2$ (White to Red)        & $+0.04$ \scriptsize{$[-0.12, +0.20]$} & $-0.02$ \scriptsize{$[-0.18, +0.14]$} & $0.63$ \scriptsize{$[0.58, 0.68]$} \\
    $1 \to 0$ (Pink to White)       & $+0.23$ \scriptsize{$[+0.06, +0.40]$} & $-0.13$ \scriptsize{$[-0.31, +0.05]$} & $0.68$ \scriptsize{$[0.61, 0.75]$} \\
    $1 \to 2$ (Pink to Red)         & $+0.29$ \scriptsize{$[+0.13, +0.45]$} & $-0.06$ \scriptsize{$[-0.23, +0.10]$} & $0.40$ \scriptsize{$[0.35, 0.44]$} \\
    $2 \to 0$ (Red to White)        & $+0.32$ \scriptsize{$[+0.16, +0.47]$} & $+0.06$ \scriptsize{$[-0.10, +0.22]$} & $0.57$ \scriptsize{$[0.51, 0.62]$} \\
    $2 \to 1$ (Red to Pink)         & $+0.29$ \scriptsize{$[+0.15, +0.44]$} & $0.00$ \scriptsize{$[-0.14, +0.15]$} & \underline{$0.39$} \scriptsize{$[0.34, 0.45]$} \\
    $0.75 \to 1$                       & \underline{$+0.35$} \scriptsize{$[+0.18, +0.54]$} & $\mathbf{+0.09}$ \scriptsize{$[-0.07, +0.26]$} & $0.64$ \scriptsize{$[0.58, 0.70]$} \\
    $0.75 \to 1.25$                    & $\mathbf{+0.38}$ \scriptsize{$[+0.20, +0.56]$} & $+0.08$ \scriptsize{$[-0.10, +0.26]$} & $0.56$ \scriptsize{$[0.51, 0.62]$} \\
    \bottomrule
  \end{tabular}}
\end{table*}

\subsection{Coloured vs. white noise}

Table~\ref{tab:main} and Fig.~\ref{fig:grid} show the comparisons between white noise with constant and scheduled coloured noise alternatives aggregated over the four tasks. An analysis of these results is provided next.

\paragraph{Learning efficiency.}

37 of the 48 coloured noise exploration strategies lead to significantly more efficient learning than white noise and none to less efficient learning. However, this advantage is task-dependent. On \reach, 44 of 48 coloured noises outperform white noise, with constant pink and red noise improving the success AUC from 0.55 to 0.67 and 0.66, respectively. On \push, only 2 coloured noises are significantly better than white noise and 2 are significantly worse. A coloured noise with $\beta=0.5$ performs best (AUC 0.26 vs.\ 0.16). On \pick, 15 of 48 coloured noises learn significantly more slowly. On \rotate, we find no significant differences with respect to white noise.

\paragraph{End performance.}

No strategy achieves significantly higher end performance than white noise. The highest aggregate end performance is achieved by a scheduled noise that increases from $\bs=0.75$ to $\be=1$. One coloured noise strategy ($0\to1.5)$ and gSDE are significantly worse than white noise. Combined with the learning efficiency outcome, these results show that coloured noise can speed up learning on some tasks without improving the success rate reached at the end of the training budget.

\paragraph{Smoothness.}

All 48 coloured schedules explore significantly more smoothly than white noise. The jerk ratio decreases from 1 for white to 0.53 for pink and 0.32 for red noise. gSDE achieves an average jerk ratio of 0.44, i.e. considerably higher than that of red noise, while both its learning efficiency and end performance are the worst overall. 
 
\subsection{When coloured noise matters most}

\begin{figure*}[!t]
    \centering
    \includegraphics[width=1\linewidth]{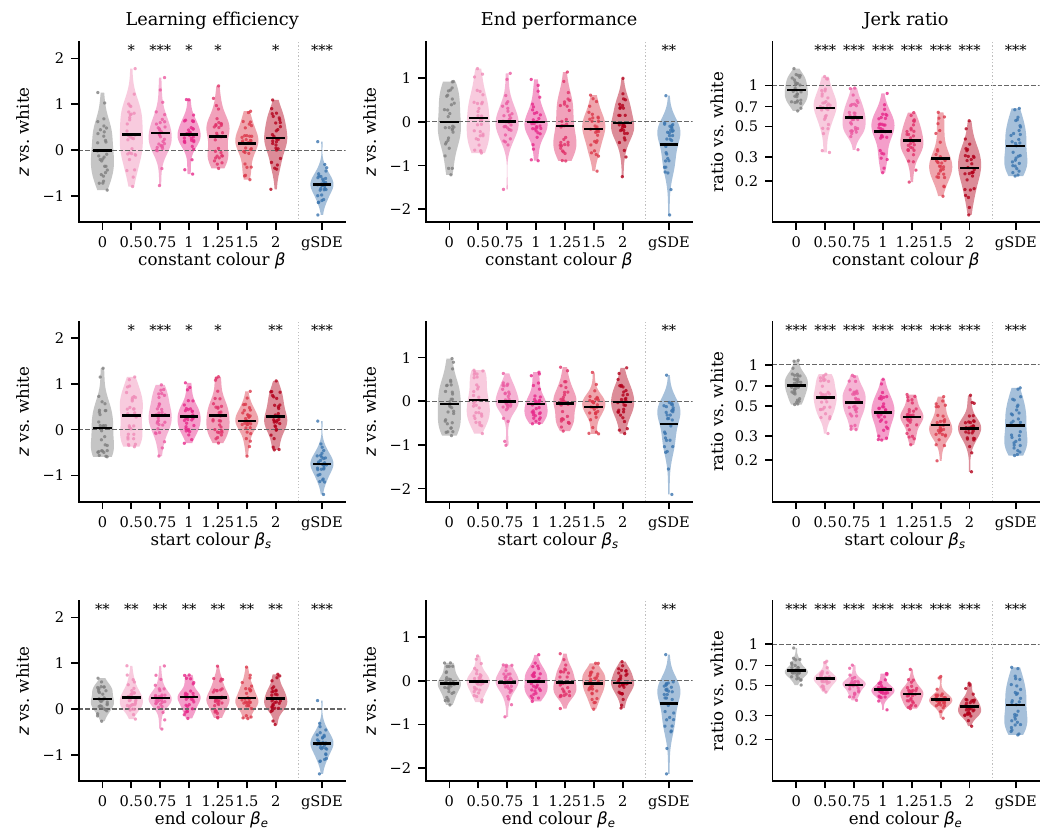}
    \caption{Effects of constant noise colour (\textbf{top}) and aggregated scheduled noise from start colour $\bs$ (\textbf{middle}) to end colour $\be$ (\textbf{bottom}) on learning efficiency (\textbf{left}), end performance (\textbf{center}), and jerk ratio (\textbf{right}). Each dot corresponds to one seed averaged over the four tasks. Horizontal black bars indicate means. Stars indicate significant differences from fixed white noise in paired Wilcoxon signed-rank tests ($^{*}p<0.05$, $^{**}p<0.01$, $^{***}p<0.001$).}
    \label{fig:violins}
\end{figure*}

To further understand the effects of coloured noise exploration during the different stages of training, we aggregate runs based on their coloured noise at the start and end of training. Results are shown in Fig.~\ref{fig:violins}.

\paragraph{End colour does not affect learning.}

Among runs that share seed and start colour, the end colour has no significant effect on learning efficiency in any task (Friedman tests, $p \geq 0.22$) or on end performance ($p \geq 0.10$). The effect is very small: ending with red rather than white noise changes the success AUC by $+0.006$ $[-0.002, 0.014]$ on \reach, $-0.001$ $[-0.019, 0.018]$ on \push, $+0.005$ $[-0.011, 0.020]$ on \pick\ and $-0.006$ $[-0.019, 0.006]$ on \rotate. Consistently, none of the 42 scheduled noises differs significantly from the constant noise of its start colour in learning efficiency or end performance (Bonferroni-corrected), and the direction of the schedule has no effect on either metric (regression coefficient of $\Delta\beta$: $-0.03$ $[-0.08, 0.01]$ for learning, $+0.00$ $[-0.03, 0.04]$ for end performance).

\paragraph{Start colour affects learning only on one task.}

Averaged over end colours, the start colour has a strong effect on \reach, where starting with more correlated noise, i.e. higher noise colour $\beta$, leads to faster learning (Friedman $p < 10^{-17}$, Spearman $\rho = 0.51$). On \push, \pick\ and \rotate, its effect is not significant ($p = 0.29$, $0.23$ and $0.41$).

\paragraph{Noise colour strongly affects smoothness.}

Ending with red rather than white noise reduces the jerk of exploration to $0.49$ $[0.47, 0.50]$ of its value on \reach, $0.50$ $[0.45, 0.55]$ on Fetch-Push, $0.60$ $[0.54, 0.67]$ on \pick\ and $0.60$ $[0.58, 0.63]$ on \rotate\ (all $p < 10^{-40}$). All 21 schedules that increase the noise colour explore significantly more smoothly than the constant noise of their start colour, at no cost in learning. At equal mean colour, jerk is slightly lower for decreasing than for increasing schedules (coefficient of $\Delta\beta$ on log jerk ratio: $+0.04$ $[0.02, 0.07]$), indicating that the colour early in training, when the policy's noise scale is largest, weighs somewhat more.
 
To summarise, our results reveal that scheduling the noise colour decouples the two consequences of coloured noise exploration. First, we find that most of the influence of the coloured noise occurs early in training, when exploration is the primary source of learning. Second, we find that the noise colour used late in training determines how smoothly the robot moves without affecting learning.

\begin{figure*}[!t]
    \centering
    \includegraphics[width=.85\linewidth]{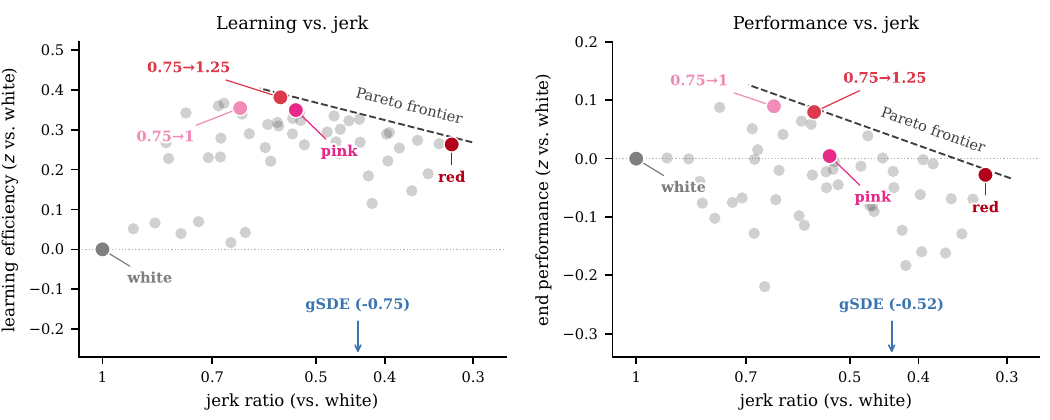}
    \caption{Trade-offs between learning efficiency (\textbf{left}) and end performance (\textbf{right}) with jerk ratio. Each point shows the aggregate from one of the 50 exploration schedules. Highlighted dots show the constant white, pink, and red noises, the two schedules with highest learning efficiency ($0.75\to1.25$) and end performance ($0.75\to1$), and gSDE. The dashed lines indicate the Pareto frontiers.}
    \label{fig:pareto}
\end{figure*}

\subsection{Balancing learning, end performance and smoothness}

Fig.~\ref{fig:pareto} shows the three metrics for all strategies and their trade-offs. The Pareto frontier of learning efficiency vs.\ jerk ratio spans nine schedules, from constant red noise (learning $z = +0.26$, jerk ratio 0.32) to the $0.75 \to 1.25$ schedule ($z = +0.38$, jerk ratio 0.56). The frontier of end performance vs.\ jerk ratio spans seven schedules, from constant red noise ($z = -0.03$, jerk ratio 0.32) to the $0.75 \to 1$ schedule ($z = +0.09$, jerk ratio 0.64). Along both frontiers, gains in learning or end performance are non-significant, whereas differences in jerk are more prominent. 39 strategies are statistically tied with the best strategy in learning efficiency, and 40 in end performance. In both sets, the strategy with the lowest jerk is constant red noise, which is the smoothest out of all 50 strategies, and is present in both Pareto frontiers.
 
Compared with white noise, red noise learns more efficiently ($\Delta z = +0.26$ $[0.04, 0.49]$), reaches the same end performance ($-0.03$ $[-0.25, 0.19]$) and explores with 3-fold less jerk (ratio 0.32 $[0.28, 0.37]$). Compared with pink noise, it does not differ in learning ($-0.08$ $[-0.29, 0.11]$) or end performance ($-0.03$ $[-0.24, 0.17]$) but explores with almost 2-fold less jerk (0.60 $[0.52, 0.68]$). Compared with the strategies at the other ends of the fronts, $0.75 \to 1.25$ and $0.75 \to 1$, red noise does not differ significantly in learning or end performance ($\Delta z$ between $-0.12$ and $-0.09$) while halving the jerk (ratios 0.57 and 0.50). Red noise also outperforms gSDE in learning ($+1.01$ $[0.81, 1.22]$), end performance ($+0.49$ $[0.25, 0.76]$) and smoothness (0.74 $[0.62, 0.86]$).

\section{Discussion}
\label{sec:discussion}

\subsection{Coloured noise supports learning with reduced movement jerk}

Across 6,000 training runs, the colour of the exploration noise had a consistent effect on how smoothly the robots explored but a smaller and task-dependent effect on how efficiently they learned. Coloured noise accelerated learning on \reach but had mixed or negligible effects on \push, \pick, and \rotate. These results contrast with the default pink noise recommendations of \cite{eberhard2023pink} based on their experiments on standard continuous control RL environments. Our sparse-reward robotic manipulation tasks with HER showed that pink noise did not learn more efficiently than other coloured noises and its advantage over white noise was moderate. This difference could be explained by specific hindrances introduced by highly correlated actions (see ~\cite{hollenstein2022action}). For example, coloured noise on the robot's gripper can keep the gripper open or closed for long periods, thus making the discovery of grasps difficult. Future work must investigate whether different coloured noises may be beneficial for different actuators within a same agent.

Interestingly, red noise was previously considered too strongly correlated as a default on standard control benchmarks~\cite{eberhard2023pink,lopez2026baby}. Two features of our experiments may explain why it performed as well as pink noise here. First, manipulation tasks often combine integrator-like dynamics (e.g. Fetch actions are Cartesian displacements of the gripper) within short episodes of 50 to 100 steps, compared with up to 1,000 steps in the DeepMind Control Suite. More correlated noise is favoured when larger displacements must be achieved within a given number of steps~\cite{eberhard2023pink}, and more strongly correlated noise moved the Fetch gripper further within an episode (Fig.~\ref{fig:fetch_noise}). Second, our low-frequency cutoff at the episode length prevents red noise from degenerating into a near-constant offset within an episode. Pink noise may therefore remain the better default for long-horizon control, whereas short-horizon manipulation benefits from the red noise smoothness.

The asymmetry we observed between the start and end colours suggests that exploration matters most early in training. As training proceeds, SAC's automatic entropy tuning reduces the policy's standard deviation $\sigma_\theta$, so that the exploration noise, whatever its colour, perturbs the agent's behaviour less. The colour of the remaining noise then shapes the motion of the robot, and hence its jerk, without necessarily changing what the agent learns.

\subsection{Smooth exploration versus smooth behaviours}

We must point out that coloured noise is not the only way to obtain smooth motions in robots. Common alternatives filter the executed actions with a low-pass filter, penalise action rates, jerk or torques in the reward function \cite{rudin2022learning}, regularise the smoothness of the learned policy \cite{mysore2021caps}, repeat actions over consecutive steps \cite{dabney2021temporally}, or simply reduce the control frequency. These methods focus on smoothing of policy. Coloured noise, in contrast, only affects exploration: the deployed policy is the deterministic mean $\tanh \mu_\theta(s_t)$, which is unaffected by the noise, so smoother training comes without deployment lag, reward modifications or additional hyperparameters. Smoothing at the level of exploration is also supported by previous evidence that temporally correlated perturbations improve state-space coverage~\cite{eberhard2023pink,hollenstein2022action,lopez2026baby} and enable learning directly on physical robots~\cite{raffin2022gsde}. The two families of methods are therefore complementary and can be combined: a smooth policy reduces the jerk of the deployed behaviour, whereas red exploration noise reduces the jerk of the exploratory behaviour that dominates during training. Among the exploration-level methods evaluated here, red noise was both smoother and considerably more efficient than gSDE.

\subsection{Limitations and future work}

This study has several limitations that need to be highlighted. All experiments were performed in simulation, and jerk was computed on the commanded actions rather than on measured on embodied motions, so the benefits of smoother exploration for physical hardware remain to be confirmed on a real robot. We considered a single algorithm (SAC with HER), linear schedules between seven noise colours, and four manipulation tasks, one of which (\reach) contributed most of the aggregate learning gains. Despite training 30 random seeds per strategy, the variability between seeds on the harder tasks was large, so that many differences in learning were not significant on individual tasks.

\section{Conclusion}
\label{sec:conclusion}

We systematically evaluated linear noise-colour schedules for exploration in robotic manipulation, covering all combinations of start and end colours between white and red noise on four tasks with a Fetch arm and a Shadow Dexterous Hand. Our results suggest a difference with respect to the standard control default pink noise recommendation of \cite{eberhard2023pink}: for off-policy robot learning, white noise should be replaced by red noise. In our experiments, fading to red resulted in exploration strategies that learned at least as efficiently as white and pink noise, reached the same end performance in aggregate, and executed the smoothest exploratory motions. Future work will focus on validating scheduled coloured noise on physical robots and explore adaptive schedules that adjust the noise colour to the learning progress of the agent.

\section*{Acknowledgements} Experiments were performed on the UNSW Compute Cluster Katana (DOI: \hyperlink{https://doi.org/10.26190/669X-A286}{10.26190/669X-A286}).

\bibliographystyle{named}
\bibliography{1_REFERENCES}

\end{document}